\documentclass[10pt,twocolumn,letterpaper]{article}

\usepackage{cvpr}
\usepackage{times}
\usepackage{epsfig}
\usepackage{graphicx}
\usepackage{amsmath}
\usepackage{amssymb}
\usepackage{algorithm}
\usepackage{algorithmic}
\usepackage{subcaption}

\usepackage[breaklinks=true,bookmarks=false]{hyperref}

\cvprfinalcopy 

\def\cvprPaperID{****} 

\ifcvprfinal\fi
\begin{document}

\title{SuperTML: Two-Dimensional Word Embedding for the Precognition on Structured Tabular Data}

\author{Baohua Sun, Lin Yang, Wenhan Zhang, Michael Lin, Patrick Dong, Charles Young, Jason Dong\\
Gyrfalcon Technology Inc.\\
1900 McCarthy Blvd. Milpitas, CA 95035\\
{\tt\small \{baohua.sun,~lin.yang,~wenhan.zhang\}@gyrfalcontech.com}
}

\maketitle

\begin{abstract}
Tabular data is the most commonly used form of data in industry according to a Kaggle ML and DS Survey. Gradient Boosting Trees, Support Vector Machine, Random Forest, and Logistic Regression are typically used for classification tasks on tabular data. DNN models using categorical embeddings are also applied in this task, but all attempts thus far have used one-dimensional embeddings. The recent work of Super Characters method using two-dimensional word embeddings achieved state-of-the-art results in text classification tasks, showcasing the promise of this new approach. In this paper, we propose the SuperTML method, which borrows the idea of Super Characters method and two-dimensional embeddings to address the problem of classification on tabular data. For each input of tabular data, the features are first projected into two-dimensional embeddings like an image, and then this image is fed into fine-tuned two-dimensional CNN models for classification. The proposed SuperTML method handles the categorical data and missing values in tabular data automatically, without any need to pre-process into numerical values. Comparisons of model performance are conducted on one of the largest and most active competitions on the Kaggle platform, as well as on the top three most popular data sets in the UCI Machine Learning Repository. Experimental results have shown that the proposed SuperTML method have achieved state-of-the-art results on both large and small datasets.
\end{abstract}

\section{Introduction}
\begin{figure*}[h]
\begin{center}
  \includegraphics[width=0.8\linewidth]{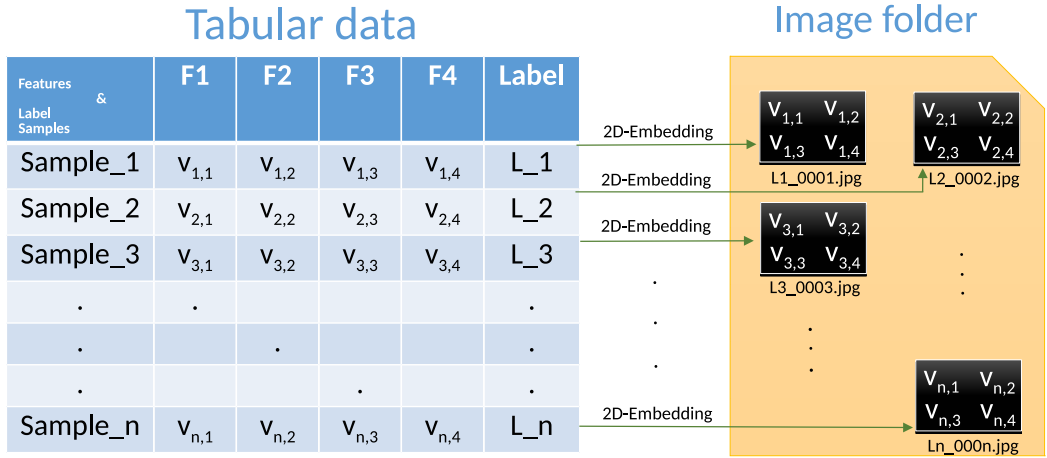}
\end{center}
\caption{An example of converting training data from tabular into images with two-dimensional embeddings of the features in the tabular data. Therefore, the problem of machine learning for tabular data is converted into an image classification problem. The later problem can use pretrained two-dimensional CNN models on ImageNet for finetuning, for example, ResNet, SE-net and PolyNet. The tabular data given in this example has n samples, with each sample having four feature columns, and one label column. For example, assume the tabular data is to predict whether tomorrow's weather is ``Sunny" or ``Rainy". The four features F1, F2, F3, and F4 are respectively ``color of the sky", ``Fahrenheit temperature", ``humidity percentage", and ``wind speed in miles per hour". Sample\_1 has class label L1=``Sunny", with four features values given by $v_{1,1} = ``blue"$, $v_{1,2} = 55, v_{1,3} = ``missing"$, and $v_{1,4} = 17$. The two-dimensional embedding of Sample\_1 will result in an image of ``Sunny\_0001.jpg" in the image folder. The four feature values are embedded into the image on different locations of the image. For example, $v_{1,1}$ is a categorical value of color $``blue"$, so the top left of the image will have exactly the alphabets of ``blue" written on it. For another example, $v_{1,2}$ is a numerical value of ``$23$", so the top right of the image will have exactly the digits of ``23" written on it. For yet another example, $v_{1,3}$ should be a numerical value but it is missing in this example, so the bottom left of the image will have exactly the alphabets of ``missing" written on it. Other ways of writing the tabular features into image are also possible. For example, ``blue" can be written in short as a single letter ``b" if it is distinctive to other possible values in its feature column. The image names will be parsed into different classes for image classification. For example, L1 = L2 = ``Sunny", and L3 = Ln =``Rainy". These will be used as class labels for training in the second step of SuperTML method.}
  \label{SuperTMLIllustration}
\end{figure*}

In data science, data is categorized into structured data and unstructured data. Structured data is also known as tabular data, and the terms will be used interchangeably. Anthony Goldbloom, the founder and CEO of Kaggle observed that winning techniques have been divided by whether the data was structured or unstructured~\cite{KaggleCEO}. Currently, DNN models are widely applied for usage on unstructured data such as image, speech, and text. According to Anthony, ``When the data is unstructured, it’s definitely CNNs and RNNs that are carrying the day"~\cite{KaggleCEO}. The successful CNN model in the ImageNet competition~\cite{ILSVRC15} has outperformed human for image classification task by ResNet~\cite{he2016deep} since 2015. And the following efforts of PolyNet~\cite{zhang2017polynet}, SE-net~\cite{hu2018squeeze}, and NASNet~\cite{zoph2018learning} keep breaking the records. Current state of the art on ImageNet given by PNASNe~\cite{liu2018progressive} achieves 82.9\% Top1 accuracy.

On the other side of the spectrum, machine learning models such as Support Vector Machine (SVM), Gradient Boosting Trees (GBT), Random Forest, and Logistic Regression, have been used to process structured data. According to a recent survey of 14,000 data scientists by Kaggle (2017), a subdivision of structured data known as relational data is reported as the most popular type of data in industry, with at least 65\% working daily with relational data. Regarding structured data competitions, Anthony says that currently XGBoost is winning practically every competition in the structured data category~\cite{KDnuggets2016}. XGBoost~\cite{chen2016xgboost} is one popular package implementing the Gradient Boosting method. Other implementations include lightgbm~\cite{ke2017lightgbm}, and catboost~\cite{prokhorenkova2018catboost}. 

Recent research has tried using one-dimensional embedding and implementing RNNs or one-dimensional CNNs to address the TML (Tabular data Machine Learning) tasks, or tasks that deal with structured data processing~\cite{lam2018neural,fastAI2018}, and also categorical embedding for tabular data with categorical features~\cite{guo2016entity,chen2016entity}. One-dimensional embeddings such as word2vec~\cite{mikolov2013distributed}, GLoVe~\cite{pennington2014glove}, fastText~\cite{joulin2016bag}, ELMO~\cite{peters2018deep}, BERT~\cite{devlin2018bert}, and Open AI GPT~\cite{radford2018improving} are widely used in NLP tasks, and as such data scientists have tried to adapt them to TML tasks. These one-dimensional embeddings project each token into a vector containing numerical values. For example, a word2vec embedding~\cite{mikolov2013distributed} could be a one-dimensional vector that acts like a 300 by 1 array.

However, this reliance upon one-dimensional embeddings may soon come to change. Recent NLP research has shown that the two-dimensional embedding of the Super Characters
method~\cite{sun2018super} is capable of achieving state-of-the-art results on large dataset benchmarks. The Super Characters method is a two-step method that was initially designed for text classification problems. In the first step, the characters of the input text are drawn onto a blank image, so that an image of the text is generated with each of its characters embedded as the pixel values in the two-dimmensional space, i.e. a matrix. The resulting image is called a Super Characters image. In the second step, Super Characters images are fed into two-dimensional CNN models for classification. The two-dimensional CNN models are trained by fine-tuning from pretrained models on large image dataset, e.g. ImageNet. This process is also known as Transfer Learning~\cite{sharif2014cnn,donahue2014decaf,yosinski2014transferable,pan2010survey}.

In this paper, we propose the SuperTML method, which borrows the concept of the Super Characters method to address TML problems. For each input, tabular features are first projected onto a two-dimensional embedding and fed into fine-tuned two-dimensional CNN models for classification. The proposed SuperTML method handles the categorical type and missing values in tabular data automatically, without need for explicit conversion into numerical type values.

Experimental results show that this proposed SuperTML method performs well on both
large and small datasets. In one instance of the Higgs Boson Machine Learning Challenge
dataset, which is ``one of the largest and most active competitions on the Kaggle platform"~\cite{chen2015higgs}, 
a single model that applied SuperTML manages to analyze 250,000 training instances
and 550,000 testing instances and obtain an AMS score of 3.979, a state-of-the-art result that
beat the previous best of 3.806~\cite{adam2015higgs}. When using the top three popular databases (ranked by number
of times accessed since 2007) from UCI Machine Learning Repository (includes the Iris dataset
(150 data instances), Adult dataset (48,482 data instances), and Wine dataset (178 data
instances)), the SuperTML method still achieved state-of-the-art results for all datasets despite
this variation in dataset size.

\section{The Proposed SuperTML Method}
The SuperTML method is motivated by the analogy between TML problems and text
classification tasks. For any sample given in tabular form, if its features are treated like stringified tokens of
data, then each sample can be represented as a concatenation of tokenized features. By applying
this paradigm of a tabular sample, the existing CNN models used in Super Characters method could be extended to be applicable to TML
problems. 


As mentioned in the introduction, the combination of two-dimensional embedding (a core
competency of the Super Characters methodology) and pre-trained CNN models has achieved
state-of-the-art results on text classification tasks. However, unlike the text classification
problems studied in~\cite{sun2018super}, tabular data has features in separate dimensions. Hence,
generated images of tabular data should reserve some gap between features in different
dimensions in order to guarantee that features will not overlap in the generated image.

SuperTML is composed of two steps, the first of which is two-dimensional embedding.
This step projects features in the tabular data onto the generated images, which will be called the
SuperTML images in this paper. The conversion of tabular training data to SuperTML image is
illustrated in Figure \ref{SuperTMLIllustration}, where a collection of samples containing four tabular features is
being sorted.

The second step is using pretrained CNN models to fine-tune on the generated SuperTML
images. 

Figure \ref{SuperTMLIllustration} only shows the generation of SuperTML images for the training data. It should
be noted that for inference, each instance of testing data goes through the same pre-processing to
generate a SuperTML image (all of which use the same configuration of two-dimensional
embedding) before getting fed into the CNN classification model.

Considering that features may have different importance for the classification task, it would be prudent to allocate larger spaces for important features and increase the font size of the corresponding feature values. This method, known as SuperTML\_VF, is described in Algorithm \ref{SuperTML_VF}.
\begin{algorithm}[tb]
\caption{SuperTML\_VF: SuperTML method with Variant Font size for embedding.}
\label{SuperTML_VF}
\textbf{Input}: Tabular data training set\\
\textbf{Parameter}: Image size of the generated SuperTML images\\
\textbf{Output}: Finetuned CNN model\\
\begin{algorithmic}[1] 
\STATE Calculate the feature importance in the given tabular data provided by other machine learning methods.
\STATE Design the location and font size of each feature in order to occupy the image size as much as possible. Make sure no overlapping among features.
\FOR{each sample in the tabular data}
\FOR{each feature of the sample}
\STATE Draw feature in the designated location and font size. 
\ENDFOR
\ENDFOR
\STATE Finetune the pretrained CNN model on ImageNet with the generated SuperTML images.
\STATE \textbf{return} the trained CNN model on the tabular data
\end{algorithmic}
\end{algorithm}

To make the SuperTML more autonomous and remove the dependency on feature
importance calculation done in Algorithm \ref{SuperTML_VF}, the SuperTML\_EF method is introduced in
Algorithm \ref{SuperTML_EF}. It allocates the same size to every feature, and thus tabular data can be directly
embedded into SuperTML images without the need for calculating feature importance. This
algorithm shows even better results than \ref{SuperTML_VF}, which will be described more in depth
later in the experimental section.
\begin{algorithm}[tb]
\caption{SuperTML\_EF: SuperTML method with Equal Font size for embedding.}
\label{SuperTML_EF}
\textbf{Input}: Tabular data training set\\
\textbf{Parameter}: Image size of the generated SuperTML images\\
\textbf{Output}: Finetuned CNN model\\
\begin{algorithmic}[1] 
\FOR{each sample in the tabular data}
\FOR{each feature of the sample}
\STATE Draw the feature in the same font size without overlapping, such that the total features of the sample will occupy the image size as much as possible.
\ENDFOR
\ENDFOR
\STATE Finetune the pretrained CNN model on ImageNet with the generated SuperTML images.
\STATE \textbf{return} the trained CNN model on the tabular data
\end{algorithmic}
\end{algorithm}

\section{Experiments}
In the experiments described below, we used the top three most popular datasets from the
UCI Machine Learning Repository~\cite{UCIMachineLearningRep} and one well-known
dataset from the Kaggle platform. These four datasets cover a variety of TML tasks. 

As of the date this paper is written, the Iris dataset~\cite{IrisData} is ranked by the UCI Machine Learning
Repository as the most popular dataset with 2.41+ million hits, followed by the Adult dataset~\cite{SalaryData} (also known as Census Salary dataset) with 1.40+ million hits and the Wine dataset~\cite{WineData} with 1.07+ million hits . Table \ref{table:DataSetsStatistics} shows the statistics of these
three datasets. The data types of the features covers a variety of integer, categorical, real, and missing values.

\begin{table*}[t]
\begin{center}
\begin{tabular}{|l|r|r|r|r|r|r|r|}
\hline 
\bf Dataset&\bf Classes&\bf \#Attributes& \bf Train& \bf Test &\bf Total& \bf Data Types&\bf Missing 
\\ \hline 
Iris&3&4& NA& NA &150& Real&No \\
Wine&3&13&NA&NA&178& Integer\& Real&No  \\
Adult&2&14&32,561&16,281&48,842& Integer \& Categorical&Yes \\
\hline 
\end{tabular}
\end{center}
\caption{Datasets statistics used in this paper from UCI Machine Learning Repository. The ``{\bf Missing}" in the table indicates whether there are missing values in the data set. The ``NA" in the table denotes that there is no given split for the training and testing dataset.}
\label{table:DataSetsStatistics}
\end{table*}

The Kaggle dataset of Higgs Boson Machine Learning Challenge is also used in the experiments. It ``attracted an unprecedented number of participants over a short period of time (May 12, 2014 to Sept 15, 2014)"~\cite{adam2015higgs}. ``There were in total 1,785 teams participating in the competition, one of the largest and most active ones on the platform website www.kaggle.com"~\cite{chen2015higgs}.  

For the second step in the SuperTML method, the
ImageNet pretrained CNN models were used in the experiments. However, a limited amount of pretrained models are publicly
available for different size of input. For example, the SE-net published pretrained model only accepts
224x224 input size using Caffe framework; while the PolyNet published model only processes inputs with size 331x331.
In order to mitigate the accuracy difference brought on by usage of different frameworks,
NASnet and PNASnet are not used because their Caffe models are not publicly available.

For all the three datasets from the UCI Machine Learning Repository, SuperTML images
of size 224x224 are generated. The pre-trained SE-net-154 model was fine-tuned on these three
datasets. We also implemented XGBoost and fine-tuned the hyper-parameters on each of the
three datasets. For Higgs Boson dataset, SuperTML both images of sizes 224x224 and 331x331
were generated for comparison of different pretrained models of SE-net-154 and PolyNet. These
two pretrained models have similar performance when working on on ImageNet (81.32\% forSE-net-154, and 81.29\% for PolyNet) but different input sizes (224x224 for SE-net-154, and 331x331 for PolyNet).

\subsection{Experiments on the Iris dataset}
``This is perhaps the best known database to be found in the pattern recognition literature"~\cite{IrisData}. The Iris dataset is widely used in machine learning courses and tutorials. It contains a total of 150 data samples, each of which represents a different subspecies of the Iris
genus of flowers (e.g. Iris Setosa, Iris Versicolor, and Iris Virginica). Each sample has a set of
four attributes and four corresponding feature values, indicating the measurements of sepal length,
sepal width, petal length, and petal width as measured in centimeters.

When implementing SuperTML to this dataset, we came across a multitude of challenges. First, the Iris dataset is very
small, with only 150 samples. If we split the training and testing 80:20, it means only there are
only 120 training samples and only 40 testing samples for each class. Deep learning models are
data hungry, and the CNN models in computer visions are especially well-known for requiring
large amounts of labeled images. For example, the ImageNet dataset has over one million images
spread relatively equally over one thousand classes. By fine-tuning on this small dataset, there is
a high tendency of overfitting. Furthermore, for this Iris dataset, the data types are all real
numbers. For methods such as Logistic Regressions, GBT, SVM, and Random Forests, the
numerical feature inputs are directly applied to the linear or non-linear models to classify the
subspecies. The CNN models used in the SuperTML method must first learn the shapes of these
numerical values and then apply nonlinear functions on the extracted image features to classify
the Iris subspecies. Just to recognize the shapes of the digits requires quite a lot of data, as shown
in MNIST dataset~\cite{deng2012mnist}.

\begin{figure}[t!]
\centering
\begin{subfigure}[t]{3.2cm}
\includegraphics[width=1\linewidth]{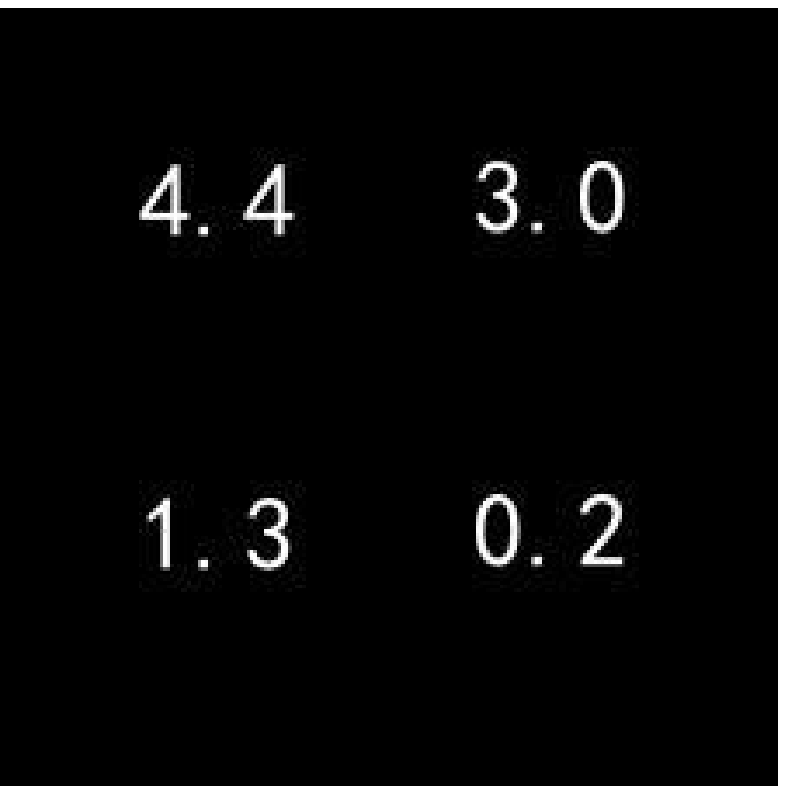}
\caption{SuperTML\_EF image example for Iris data. Each feature is given equal importance in this example.}\label{SuperTMLexampleImageIris}
\end{subfigure}\qquad
\begin{subfigure}[t]{3.2cm}
\includegraphics[width=1\linewidth]{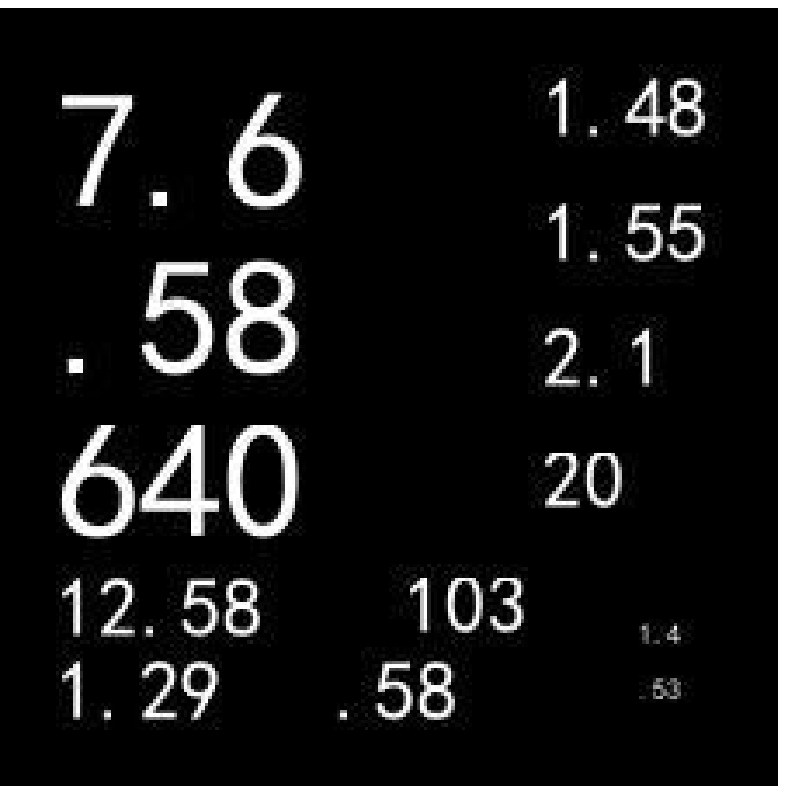}
\caption{SuperTML\_VF image example for Wine data. Features are given different sizes according to their importance.}\label{SuperTMLexampleImageWine}
\end{subfigure}
\caption{Examples of generated SuperTML image for Iris and Wine dataset.}
\end{figure}

Figure \ref{SuperTMLexampleImageIris} shows an example of a generated SuperTML image, created using Iris data.
Different from existing methods that first convert every feature value into numerical type, as typically done for preprocessing when using XGboost, SVM and etc., the idea of converting tabular features of any data type into string type may seem to increase the difficulty for training model and inference. However, this proposed counter-intuitive method of SuperTML outperforms existing methods as shown in the experimental results. The experimental results of using SE-net-154 shown in Table \ref{UCIresults} is based on an 80:20 split of
the 150 samples. It shows that the proposed SuperTML method achieves the same accuracy as
XGBoost on this small dataset.
\begin{table*}[t]
\begin{center}
\begin{tabular}{|l|r|r|r|}
 \hline 
\bf Accuracy &\bf Iris(\%) &\bf Wine(\%) & \bf Adult(\%) \\
 \hline
xgboost &93.33 &96.88 & 87.32 \\
GB~\cite{biau2018accelerated} & -- & --& 86.20  
\\ \hline
SuperTML & \bf93.33  &  \bf97.30  & \bf87.64\\\hline
\end{tabular}
\caption{Model accuracy comparison on the tabular data from UCI Machine Learning Repository. The splits on Iris and Wine data is 80\%:20\% as described in the experimental setup.}\label{UCIresults}
\end{center}
\end{table*}
\subsection{Experiments on the Wine dataset}
The Wine dataset shares a few similarities to the Iris dataset, so we conducted this
experiment immediately after the Iris experiment. This dataset has the similar task of classifying
the input samples into one of the three classes and comprises of only 178 samples, making it also
a small dataset. The input features are measurements on alcohol, hue, ash, and etc.. In addition,
there is no given split of training and testing datasets in this Wine dataset, another similarity
between it and the Iris dataset. The number of attributes is 13, which is more than 4 times of that
of the Iris dataset. In this set, the features’ data types includes not just real numbers, but also
integers. These differences make the classification on Wine data with SuperTML method even
harder than in the Iris dataset for the SuperTML image because of the increased number of
features and variation in space due to different data types.

For this dataset, we use SuperTML VF, which gives features different sizes on the
SupterTML image according to their importance score. The feature importance score is obtained
using the XGBoost package~\cite{chen2016xgboost}. One example of a SuperTML image
created using data from this dataset is shown in Figure \ref{SuperTMLexampleImageWine}. The importance score shows that the
feature of color intensity is the most important, so we allocate font size of 48 to it in the 224x224
image (can be seen in the top right corner). The following features of importance are flavanoids
and proline, which were allocated space on the left and given font size 48. This pattern of
importance and font size is applied to all features, all the way down to the least important
features of proanthocyanins and nonflavanoid phenols, which were placed in the bottom right corner and given a font size of 8. The results in Table \ref{UCIresults} shows that the SuperTML method obtained a slightly better accuracy than XGBoost on this dataset.
\subsection{Experiments on the Adult dataset}
\label{SalaryData}
The task of this Adult dataset is to predict whether a person’s income is larger or smaller
than 50,000 dollars per year based on a collection of surveyed data. Each sample of data (each
person) has 14 attributes, including age, gender, education, and etc.. These attributes are stored
using a combination of integer, real numbers, and categorical data. It has 32,561 training samples
and 16,281 testing samples. Compared with the other two datasets from the UCI Machine
Learning Repository, this relatively large dataset is in favor of deep learning models that
implement the SuperTML method.

\begin{figure}[t!]
\begin{center}
  \includegraphics[width=0.5\linewidth]{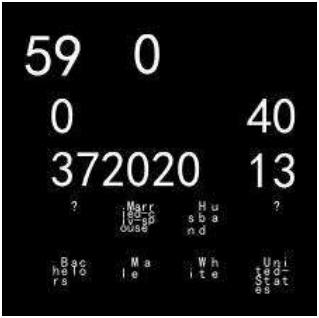}
  \caption{SuperTML image example from Adult dataset. This sample has yearly salary larger than 50k with its features given different sizes in the SuperTML image according to their importance given by third party softwares. This sample has age = 59, capital gain = 0, capital loss = 0, hours per week = 40, fnlweight = 372020, education number = 13, occupation = "?" (missing value as given in the data, it can also be replaced by "MissinngValue" in the SuperTML image), marital status = "Married-civ-spouse", relationship = "Husband", workclass = "?" (missing value), education = "Bachelors", sex = "Male", race = "White", native country = "United-States".}\label{SuperTMLexampleImageSalaryWithMissingValue}
\end{center}
\end{figure}

For categorical features that are represented by strings, the Squared English Word (SEW)
method~\cite{sun2019squared} is used. The benefits of using the English word in this format is
two-folded. Firstly, the Super Characters method has shown state-of-the-art performance when
processing Asian languages, which has their characters written in the form of glyphs enclosed in
a squared space. Building of Super Characters, SEW converts each English word into a square
comprised of its characters and guarantees that each word will be written in a unique way.
Secondly, writing the features expressed by English strings in this format guarantees that each
feature occupies the same position without any change caused by the length of the feature string.
One example of a generated SuperTML image is given in Figure \ref{SuperTMLexampleImageSalaryWithMissingValue}.
Table \ref{UCIresults} shows the results on Adult dataset. On this dataset, Biau and et. a.l.~\cite{biau2018accelerated} proposed an Accelerated Gradient
Boosting (AGB) model and compared the performance with an original Gradient Boosting (GB) model using a series of fine-tuned hyper-parameters. The best accuracy is given by the GB model when the shrinkage parameter is set at 0.1. We also tried implementing XGBoost on thisdataset and preprocessed the categorical data by using integer encoding (using the Python pandas library with astype(`category')).The XGBoost’s best result was 87.32\% accuracy after
fine-tuning the number of trees at 48. We can see that on this dataset, the SuperTML method still
has a higher accuracy than the fine-tuned XGBoost model, outperforming it by 0.32\% points of accuracy.
\subsection{Experiments on the Higgs Boson Machine Learning Challenge dataset}
\label{HiggsBossonExperiments}
The Higgs Boson Machine Learning Challenge involved a binary classification task to
classify quantum events as signal or background. It was hosted by Kaggle, and though the
contest is over, the challenge data is available on opendata~\cite{adam2015higgs}. It has
25,000 training samples, and 55,000 testing samples. Each example has 30 features, each of
which is stored as a real number value. In this challenge, AMS score~\cite{adam2014learning} is used as the performance metric.

The reason why we selected this dataset is two-fold. First, it is a well-known dataset and
successful models such as XGBoost and Regularized Greedy Forest have been used in this dataset.
Second, the performance metric used in this dataset is AMS score instead of accuracy. It is useful
to test the performance of SuperTML method using a different metric and compare its results to
other leading options.

\begin{figure}[t!]
\centering
\begin{subfigure}[t]{3.2cm}
\includegraphics[width=1\linewidth]{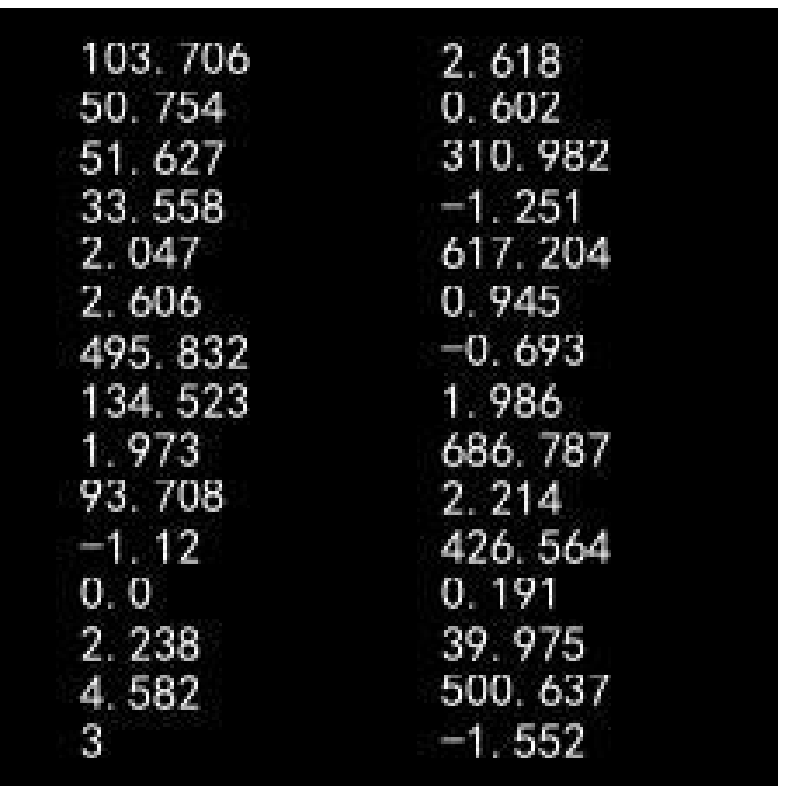}
\caption{SuperTML image example for Higgs Boson data. Each feature is given equal importance in this example.}\label{SuperTMLexampleImageHiggsBosonEqual}
\end{subfigure}\qquad
\begin{subfigure}[t]{3.2cm}
\includegraphics[width=1\linewidth]{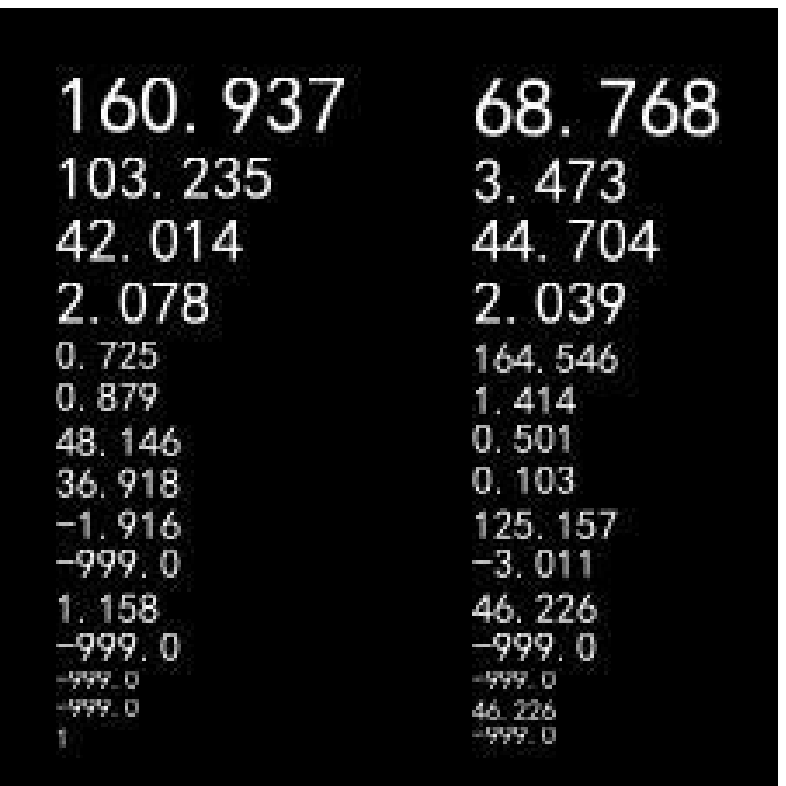}
\caption{SuperTML image example for Higgs Boson data. Features are given different sizes according to their importance.}\label{SuperTMLexampleImageHiggsBosonNonEqual}
\end{subfigure}
\caption{Examples of generated SuperTML image for Higgs Boson dataset.}
\end{figure}

The SuperTML images of size 224x224 are generated for fine-tuning the SE-net models,
and the images of size 331x331 are generated for fine-tuning the PolyNet models. Figure \ref{SuperTMLexampleImageHiggsBosonEqual}
shows an example of example of a “background” event, which is generated into an SuperTML
image of 224x224 through a SuperTML\_EF method. Figure 4b shows an example of a “signal”
event, generated through a SuperTML\_VF method, which also in an 224x224 image. The
331x331 SuperTML images are similar to 224x224 images except that each feature’s font size
and allocated space is increased.

As pointed out by~\cite{chen2016xgboost}, the AMS is an unstable measure, and AMS
was not chosen as a direct objective for XGBoost implementation in this challenge. 
For simplicity, cross-entropy loss is still used in this dataset as an objective to minimize. Table \ref{HiggsBosonTable} shows the
comparison of different algorithms. The DNN method and neural networks used in the first and
third rows are using the numerical values of the features as input to the models, which is
different from the SuperTML method of using two-dimensional embeddings. It shows that
SuperTML\_EF method gives the best AMS score of 3.979. The PolyNet models trained with
larger size of 331x331 does not help improve the AMS score. In addition, the SuperTML\_EF
gives better results than SuperTME\_VF results for both 224x224 and 331x331 image sizes,
which indicates SuperTML method can work well without the calculation of the importance
scores.

\begin{table}[t]
\begin{center}
\begin{tabular}{|l|r|}
\hline\bf Methods   & \bf AMS      \\
\hline
DNN by Gabor Meli   & 3.806      \\
RGF and meta ensemble   & 3.789      \\
Ensemble of neural networks    & 3.787      \\
XGBoost & 3.761      
\\ \hline
SuperTML\_EF(224x224)   & \bf3.979      \\
SuperTML\_VF (224x224)   & 3.838      \\
SuperTML\_EF (331x331)      & 3.934      \\
SuperTML\_VF (331x331)   & 3.812      \\
\hline
\end{tabular}
\caption{Comparison of AMS score on Higgs Boson dataset for different methods. The first four rows are top rankers in the leaderboard in the Higgs Boson Challenge.}\label{HiggsBosonTable}
\end{center}
\end{table}

\begin{figure}[t!]
\centering
\begin{subfigure}{3cm}
\includegraphics[width=1\linewidth]{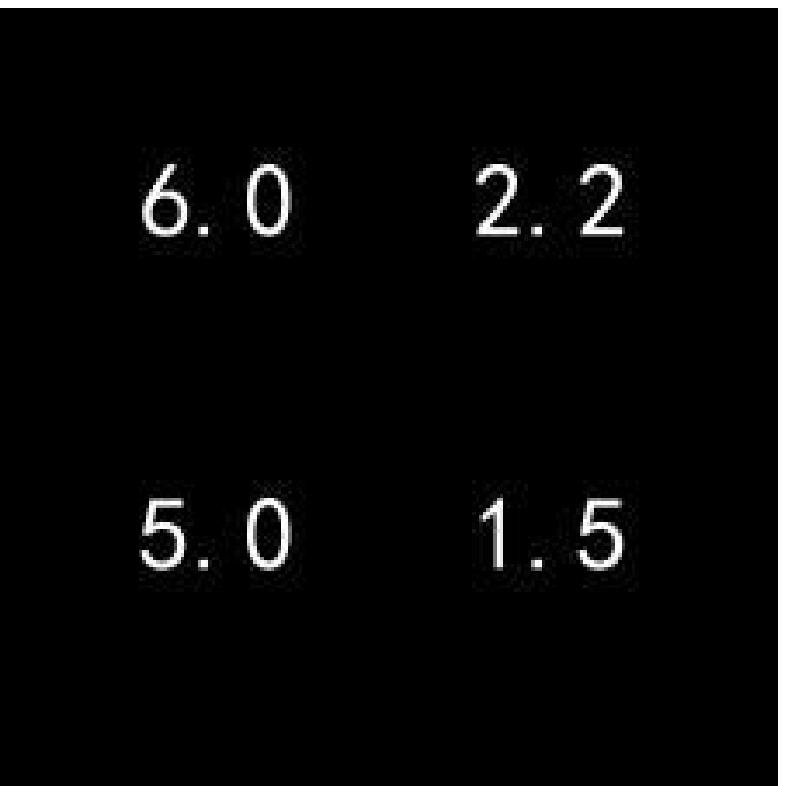}
\caption{The test SuperTML image of a virginica sample.}\label{ErrorAnalysis1}
\end{subfigure}\qquad
\begin{subfigure}{3cm}
\includegraphics[width=1\linewidth]{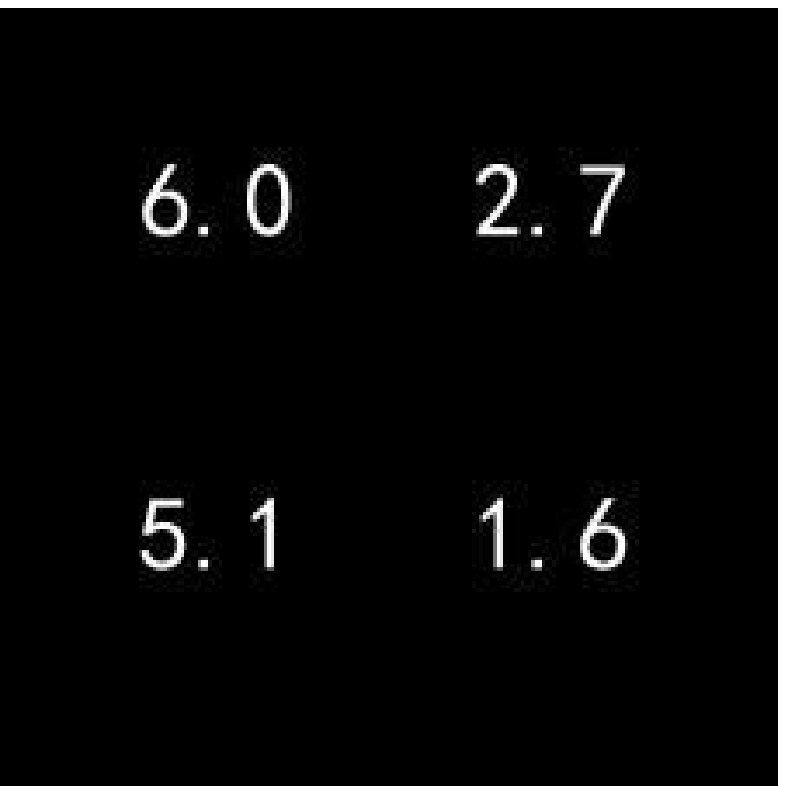}
\caption{One example from training set with label ``versicolor".}\label{ErrorAnalysis2}
\end{subfigure}
\\
\begin{subfigure}{3cm}
\includegraphics[width=1\linewidth]{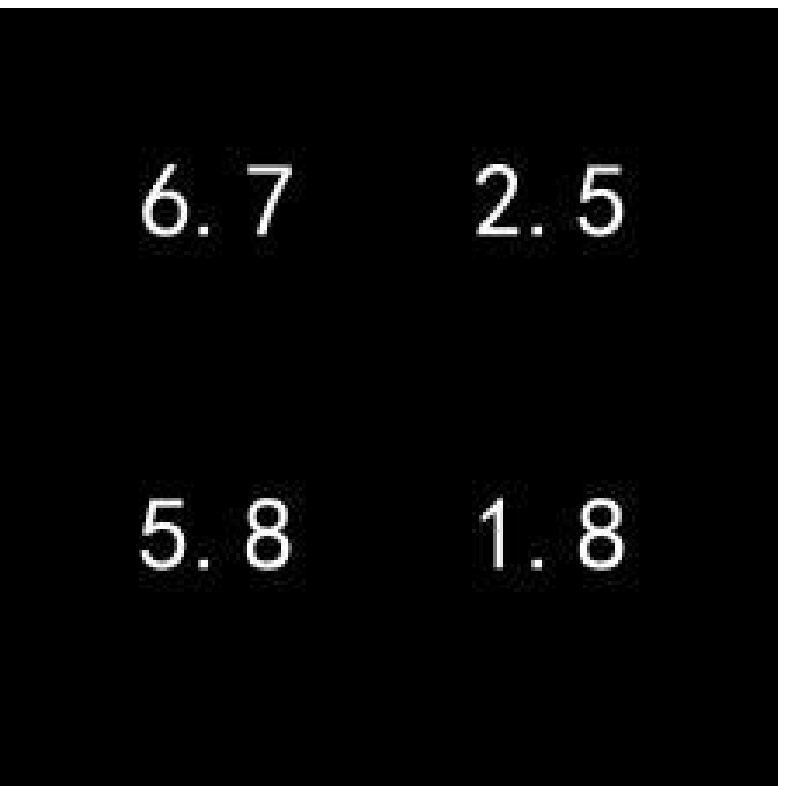}
\caption{First example from training set with label ``virginica".}\label{ErrorAnalysis3}
\end{subfigure}\qquad
\begin{subfigure}{3cm}
\includegraphics[width=1\linewidth]{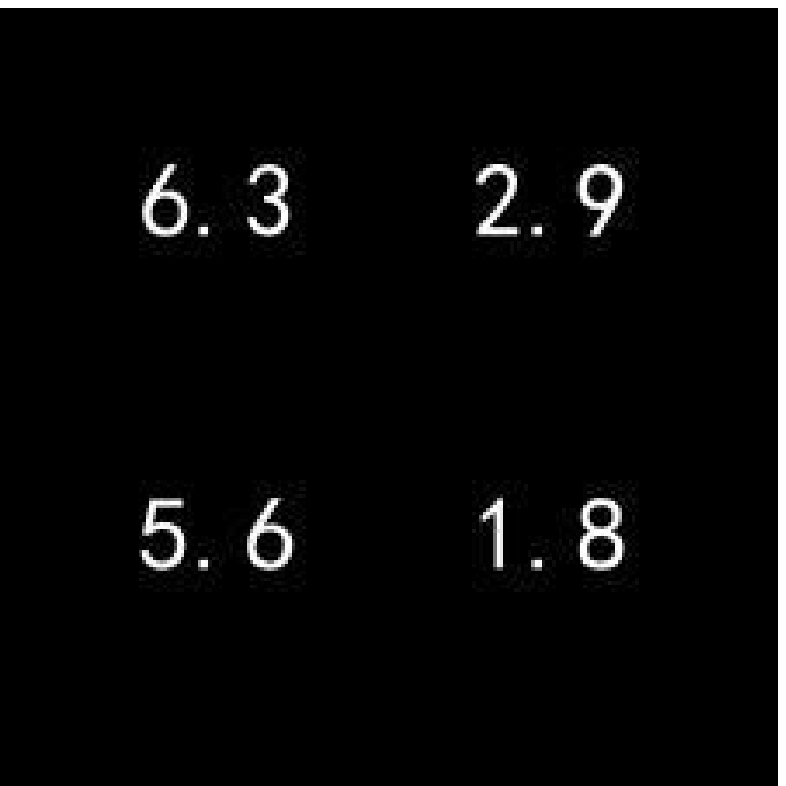}
\caption{Second example from training set with label ``virginica".}\label{ErrorAnalysis4}
\end{subfigure}
\\
\begin{subfigure}{3cm}
\includegraphics[width=1\linewidth]{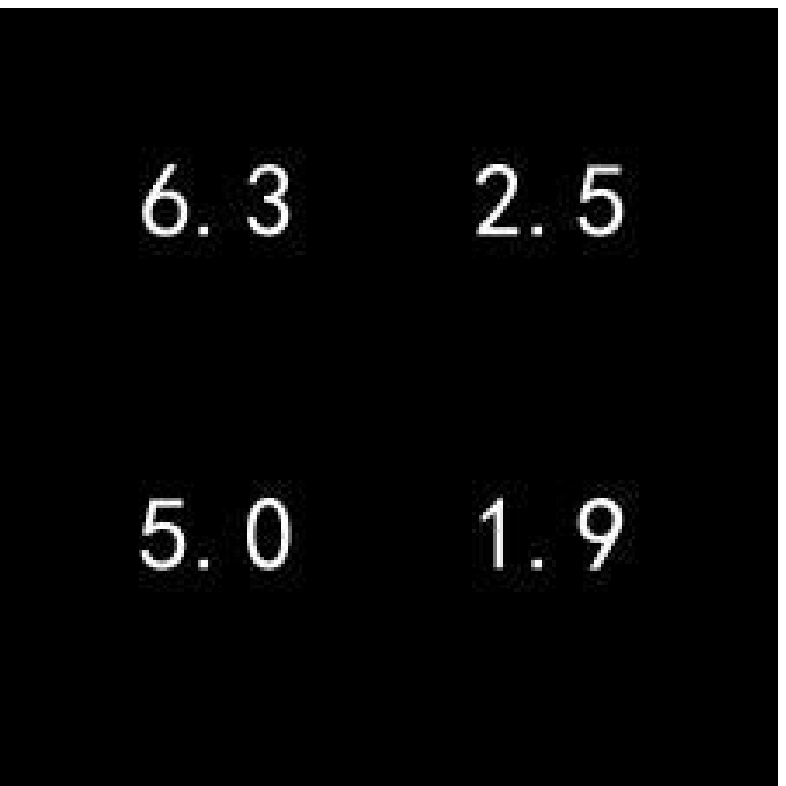}
\caption{Third example from training set with label ``virginica".}\label{ErrorAnalysis5}
\end{subfigure}\qquad
\begin{subfigure}{3cm}
\includegraphics[width=1\linewidth]{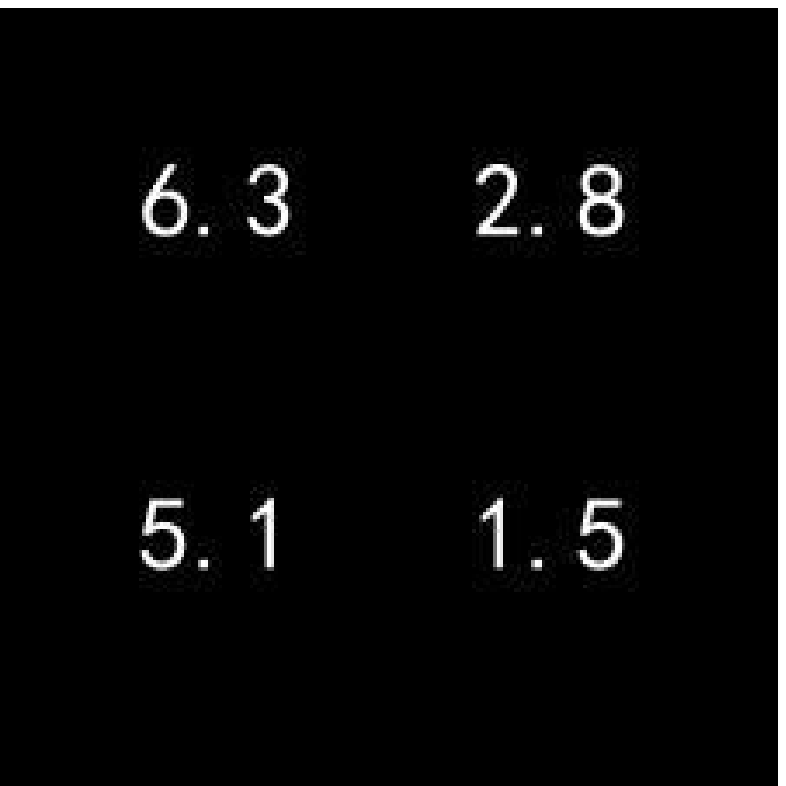}
\caption{Fourth example from training set with label ``virginica".}\label{ErrorAnalysis6}
\end{subfigure}
\caption{Error analysis on an Iris-viginica input wrongly predicted as Iris-versicolor. These six samples have the common integer for each of the four feature values, namingly 6, 2, 5, and 1. But these samples may have different decimal values. Figure \ref{ErrorAnalysis3}-Figure \ref{ErrorAnalysis6} are the only four training samples with the integer portion of the feature values same as Figure \ref{ErrorAnalysis1}. But one ``versicolor" sample from training set as shown in Figure \ref{ErrorAnalysis2} not only has the same integer part as Figure \ref{ErrorAnalysis1}, but also has more similar shapes of decimal part for each of the four feature values to Figure \ref{ErrorAnalysis1} than the other four samples. The CNN models that learn the classification model based on the shape of the feature values written on the image have high tendency to classify the example in Figure \ref{ErrorAnalysis1} as the same category of Figure \ref{ErrorAnalysis2}.}
\end{figure}

\subsection{Error Analysis}
The experiment using the SuperTML method on the Iris dataset with 80:20 split for
training and testing had 2 incorrect predictions. We will be taking one of the wrongly predicted
sample in the testing dataset for error analysis. Its ground truth label is Iris-virginica, but was
incorrectly classified as Iris-versicolor. Its four features are 6.0, 2.2, 5.0, and 1.5 respectively, as
shown in Figure \ref{ErrorAnalysis1}. All the training samples with common integer portion for each of the four
feature values, namely 6, 2, 5, and 1 are taken into comparison in Figure \ref{ErrorAnalysis2}-Figure \ref{ErrorAnalysis6}. However,
these samples may have different decimal values. It shows that this SuperTML image of this
virginica example in Figure \ref{ErrorAnalysis1} looks more like the versicolor sample in Figure Figure \ref{ErrorAnalysis2} than the other
virginica samples in Figure \ref{ErrorAnalysis3}-Figure \ref{ErrorAnalysis6}, when the shape of numbers in decimal portion is
compared. Hence, the testing sample of virginica in Figure \ref{ErrorAnalysis1} is more likely to be classified as
versicolor, which is the label for Figure \ref{ErrorAnalysis2}.

The features in this Iris dataset are all numerical values without missing numbers. During
model training, these SuperTML images of numerical features are fed into the two-dimensional
CNN model which classifies images based on the pixel values and the relationship between
pixels. At inference time, the model classifies the samples based on the “appearance” of the
features in real-valued numbers. However, the numerical values have some hidden relationship
behind the shape of the digits, such as 6.01 and 5.999. They both approximate to the number 6.00
even though their shape is not alike. This is hard for the CNN model to learn.

\section{Conclusion and Future Work}
The proposed SuperTML method borrows the idea from Super Characters and
two-dimensional embedding and fine-tunes the pre-trained CNN models on unstructured data for
transfer learning the structured data in the tabular form. Experimental results shows that the
proposed SuperTML method has achieved state-of-the-art results on both large and small tabular
dataset. As low power domain specific CNN accelerators~\cite{sun2018ultra,sun2018mram} become available in
the market, the SuperTML method can realize its huge potential for practical applications in the
real world. For example, for IoT (Internet of Things) applications in smart homes, current
machine learning solutions implemented at the edge level are still using Logistic Regression
models. These regression models are computationally inexpensive but are expected to be much
less accurate when compared to large models like CNN. Using these low-power CNN
accelerators with the SuperTML method, it will become possible to provide low-power and high
accuracy models at the edge devices. The future work is projected to go in four directions. First,
given the success of Super Characters method~\cite{sun2018super,sun2019squared} in text classification,
categorical type of data, and also missing value calculations, the SuperTML method should be
able to be directly supplement or even replace the current models in that field. Unlike numerical
features, the categorical feature has all the information written in the text. Second, compared
with not only the Gradient Boosting method, but also the one-dimensional embedding based
RNN and CNN methods, SuperTML could become the new state-of-the-art in computing and
solving TML tasks. Third, the SuperTML method could be enlarged to provide support for more
powerful CNNs such as NASNet, PNASnet, and the others. Fourth, by modifying and improving the model architectures, variant feature importance calculations could be improved in order to find a more accurate way to implement attention~\cite{bahdanau2014neural,vaswani2017attention} scheme.

{\small
\bibliographystyle{ieee_fullname}
\bibliography{egbib}
}

\end{document}